\documentclass[letterpaper, 10 pt, conference]{ieeeconf}  % Comment this line out if you need a4paper

\IEEEoverridecommandlockouts                              % This command is only needed if 
\usepackage{cite}
\usepackage{amsmath,amssymb,amsfonts}
\usepackage{algorithmic}
\usepackage{graphicx}
\usepackage{textcomp}
\usepackage{xcolor}
\usepackage{enumerate}
\usepackage[linesnumbered,ruled,lined]{algorithm2e}
\usepackage{booktabs}
\usepackage{threeparttable}
\usepackage{multirow}
\usepackage{stfloats}
\usepackage{subfigure}
\usepackage{graphicx}
\usepackage{rotating}
\usepackage{geometry}
\usepackage{hyperref}
\title{\LARGE \bf
PAS-SLAM: A Visual SLAM System for Planar Ambiguous Scenes
}

\author{Xinggang Hu$^{1,3}$, Yanmin Wu$^{2}$, Mingyuan Zhao$^{3}$, Linghao Yang$^{4}$, Xiangkui Zhang$^{1}$, Xiangyang Ji$^{3*}$ % <-this % stops a space
	\thanks{$^{1}$School of Electronic Information and Electrical Engineering, Dalian University of Technology, Dalian, China.}%
	\thanks{$^{2}$School of Electronic and Computer Engineering, Peking University, Shenzhen, China.}%
	\thanks{$^{3}$Department of Automation, Tsinghua University, Beijing, China.}%
	\thanks{$^{4}$College of Information Science and Engineering, Northeastern University, Shenyang, China.}%
	\thanks{$^*$Corresponding author. (Email: {\tt\small 
xyji@tsinghua.edu.cn}) }
}

\begin{document}

\maketitle
\thispagestyle{empty}
\pagestyle{empty}

%%%%%%%%%%%%%%%%%%%%%%%%%%%%%%%%%%%%%%%%%%%%%%%%%%%%%%%%%%%%%%%%%%%%%%%%%%%%%%%%
% 环境中的动态因素会因违背SLAM算法的静态环境假设导致相机定位精度下降。
\begin{abstract}
% 基于平面特征的视觉SLAM在环境结构感知、增强现实等领域中有着广泛应用。但是目前的相关研究面临着在复杂场景下定位和建图表现不佳的问题，这主要是所使用的平面特征以及数据关联方法准确性较差引起的。在本文中，我们提出了一种面向复杂场景的基于平面特征的视觉SLAM系统，包含平面处理、数据关联和多约束因子图优化。我们提出了一种平面处理策略，将语义信息与平面结合，并提取平面的边缘线和顶点，以应用于平面筛选、数据关联和位姿优化等环节；然后提出了一种集成数据关联策略，结合了平面参数、语义信息、投影IoU以及非参数检验，实现了复杂场景下准确、稳健的平面数据关联；最后设计了一套多种约束因子图进行相机位姿优化。使用公共数据集进行的定性和定量实验表明，与最先进的方法相比，所提出的系统在地图构建和相机定位的准确性和鲁棒性方面达到了有力的竞争水平。
Visual SLAM (Simultaneous Localization and Mapping) based on planar features has found widespread applications in fields such as environmental structure perception and augmented reality. However, current research faces challenges in accurately localizing and mapping in planar ambiguous scenes, primarily due to the poor accuracy of the employed planar features and data association methods. In this paper, we propose a visual SLAM system based on planar features designed for planar ambiguous scenes, encompassing planar processing, data association, and multi-constraint factor graph optimization. We introduce a planar processing strategy that integrates semantic information with planar features, extracting the edges and vertices of planes to be utilized in tasks such as plane selection, data association, and pose optimization. Next, we present an integrated data association strategy that combines plane parameters, semantic information, projection IoU (Intersection over Union), and non-parametric tests, achieving accurate and robust plane data association in planar ambiguous scenes. Finally, we design a set of multi-constraint factor graphs for camera pose optimization. Qualitative and quantitative experiments conducted on publicly available datasets demonstrate that our proposed system competes effectively in both accuracy and robustness in terms of map construction and camera localization compared to state-of-the-art methods.
\end{abstract}

%%%%%%%%%%%%%%%%%%%%%%%%%%%%%%%%%%%%%%%%%%%%%%%%%%%%%%%%%%%%%%%%%%%%%%%%%%%%%%%%
\section{INTRODUCTION}
% 由于视觉相机体积和重量小便于携带、成本低廉以及能够从观察到的场景中得到大量的信息，视觉SLAM在机器人导航  、自动驾驶  和增强现实   等领域得到了广泛应用。基于特征的视觉SLAM方法从图像中提取点、线、面等特征，并确定特征之间的对应关系以估计相机位姿。基于点特征的视觉SLAM方案发展较早且涌现了很多经典的研究工作      。基于点特征的视觉SLAM系统在结构单一、纹理缺失的环境中容易跟踪失，相比之下，环境中的面特征作为室内环境中稳定的物理结构具有较强的鲁棒性、长时间观测的稳定性，所以近年来有很多研究工作[]将面特征加入到视觉SLAM中。
Due to the low cost of visual cameras and their ability to capture a wealth of information from the observed scene, visual SLAM has found extensive applications in fields such as robotics navigation\cite{yuan2019multisensor,chiang2019seamless}, autonomous driving\cite{henein2020dynamic,fernandes2021point}, and augmented reality\cite{jinyu2019survey,wu2023object,park2022strategy}. Feature-based visual SLAM methods extract points, lines, and planar features from images and establish correspondences between these features to estimate camera poses. Point-feature-based visual SLAM solutions have a long history and have spawned numerous classical research endeavors\cite{davison2007monoslam,klein2007parallel,mur2015orb,mur2017orb,sumikura2019openvslam,campos2021orb}. However, point-feature-based visual SLAM systems tend to lose track in environments with a single structure or lacking texture. In contrast, planar features within the environment, as stable physical structures in indoor settings, exhibit strong robustness and long-term stability. Consequently, recent years have witnessed substantial research\cite{zhang2019point,li2020co,wang2021tt,shu2023structure} efforts aimed at incorporating planar features into visual SLAM.

\begin{figure}[]
% 	\vspace{-3mm}
\centering
 	\includegraphics[scale=0.32]{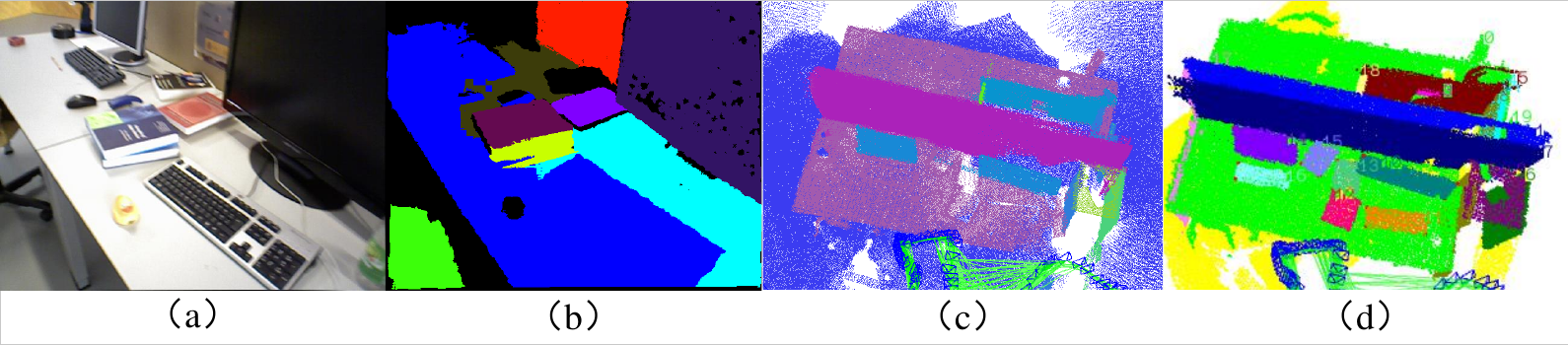}
 	\setlength{\abovecaptionskip}{-0.4cm} % 控制标题与图之间的距离
  % Overview。（a）和（b）平面模糊场景；（c）当前相关研究难以做到准确建图；（d）我们方法的建图效果；
 	\caption{Overview: (a) and (b) show scenes with plane ambiguity; (c) current related research struggles to achieve accurate mapping; (d) mapping results of our approach.}
 	\label{Overview}
 	\vspace{-7.5mm} % 控制整个图（包括标题）与下面的距离
 \end{figure}

% 目前的以平面为特征的视觉SLAM系统主要解决在纹理缺失环境下的基于点特征的视觉SLAM定位精度较差甚至定位失败的问题，面向的场景的平面结构一般比较简单。但是现实中也存在很多平面混淆场景且面向此类场景的平面SLAM的研究具有重要的意义，例如室内环境中的桌面上一般会摆放很多物体，如图1（a）所示，这些物体一方面会影响桌面平面的构建；另一方面有些形状规则的物体上也可以提取到平面，但是这些平面与从桌面上提取到的平面参数非常相似、重叠度很高，难以分辨。将桌面和这些形状规则的物体上提取到的平面准确构建出来具有重要的价值：①可以提高对环境的结构感知能力；②可以为相机位姿估计、物体位姿估计等问题提供更丰富、准确的约束；③可以服务于增强现实，为虚拟物体注册提供准确的平面锚点。
% 目前的以平面为特征的视觉SLAM系统主要解决在纹理缺失环境下的基于点特征的视觉SLAM定位精度较差甚至定位失败的问题，面向的场景的平面结构一般比较简单。然而，实际应用中经常面临具有复杂平面混淆的场景。以室内环境为例，在桌面上通常存在多个物体，如图1（a）所示，这些物体不仅对桌面平面的建模构成了挑战，还存在某些形状规则的物体上也能提取到平面。但是这些平面与桌面上提取到的平面参数相似度和重叠度极高，难以分辨，如图1（b）所示。。研究面向此类场景的平面SLAM具有重要的理论和实际意义，例如精确地重建桌面以及从这些形状规则的物体上提取的平面具有以下重要价值：①提高对环境的结构感知能力；②为相机位姿估计、物体位姿估计等问题提供更丰富、准确的约束；③服务于增强现实，为虚拟物体注册提供准确的平面锚点。
% 目前相关工作在这种复杂场景下的表现一般较差，难以构建准确的平面地图，如图1（c）所示，且平面特征的增加可能会对相机位姿估计产生负面影响，这主要是由于以下原因导致的：①平面提取很难兼顾丰富、准确和稳定，如图1（b）所示，而且缺乏有效的平面处理机制；②平面特征缺乏描述子，目前相关研究的平面特征关联策略比较简单，在复杂环境下由于有些平面的参数非常相近导致这些策略极易失效。
Presently, visual SLAM systems that rely on planar features primarily address the issue of poor localization accuracy or even failure in point-feature-based visual SLAM in texture-deprived environments. The planar structures of the scenes addressed in these works are generally relatively simple. However, planar ambiguous scenes are frequently encountered in real applications. Taking indoor environments as an example, tabletops often host numerous objects, as shown in Fig.\ref{Overview}(a). These objects not only pose challenges for modeling the desktop plane, but planes can also be extracted from some regularly shaped objects. However, the similarity in parameters and the degree of overlap between planes extracted from regular objects and those from the tabletop are both extremely high, making them difficult to distinguish, as illustrated in Fig.\ref{Overview}(b). Studying planar SLAM oriented towards such scenes has important theoretical and practical significance: 1) Enhanced environmental structural perception; 2) Provision of richer and more accurate constraints for tasks such as camera pose estimation and object pose estimation; 3) Support for augmented reality applications by furnishing precise plane anchors for the registration of virtual objects. Currently, related research tends to underperform in such planar ambiguous scenes, struggling to construct accurate plane maps, as shown in Fig.\ref{Overview}(c). Additionally, the addition of planar features may negatively impact camera pose estimation. This is primarily due to the following reasons: (1) Plane extraction is challenging to balance richness, accuracy, and stability, as illustrated in Fig.\ref{Overview}(b), and lacks effective plane processing mechanisms; (2) Planar features often lack descriptors, and the current research's strategies for planar feature association are relatively simplistic. In planar ambiguous scenes, these strategies can easily fail.

% ---->>>>>>>>>>>>> 图1 系统概述

\begin{figure*}[]
% 	\vspace{-3mm}
\centering
 	\includegraphics[scale=0.52]{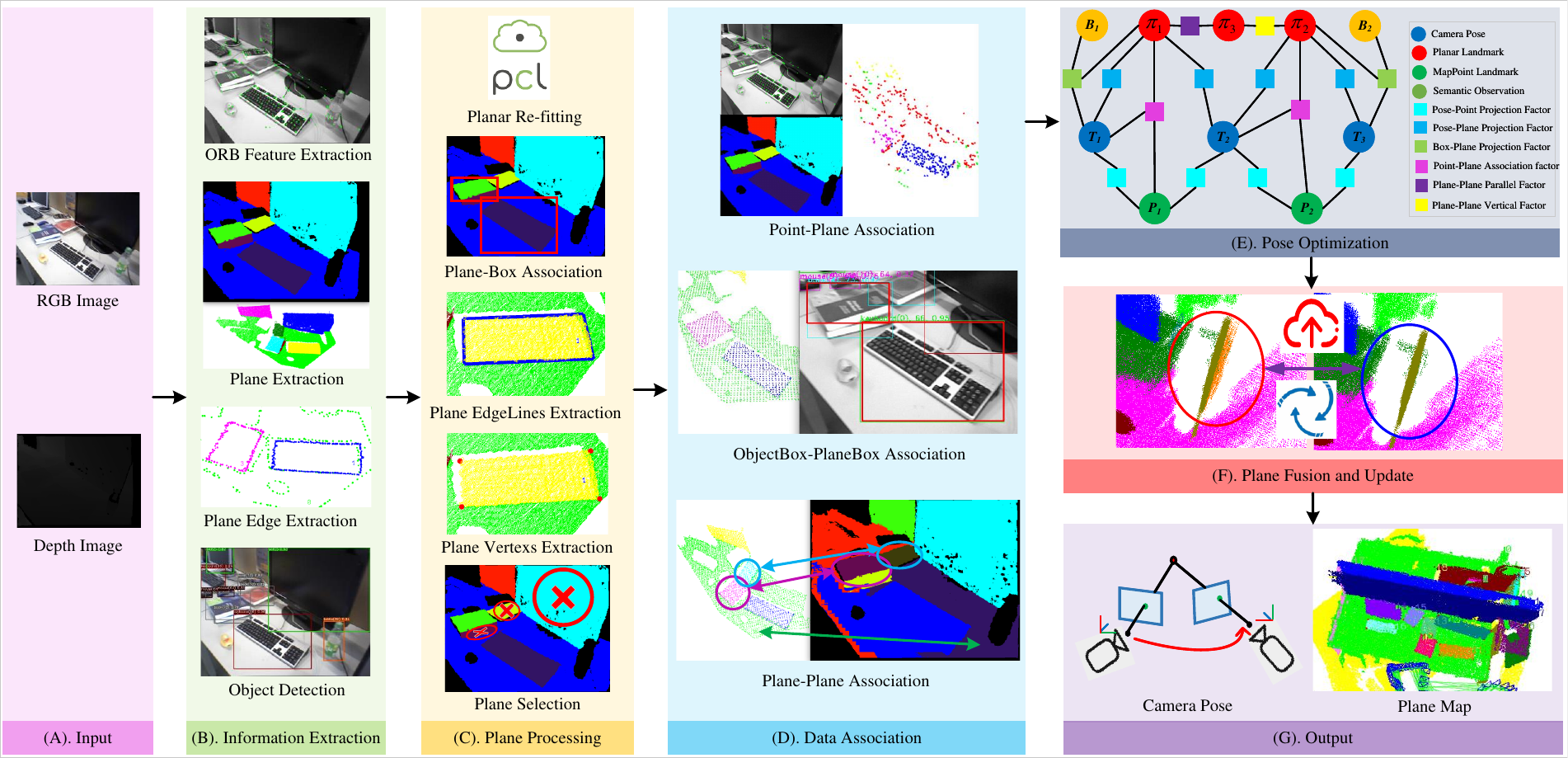}
 	\setlength{\abovecaptionskip}{-0.1cm} % 控制标题与图之间的距离
 	\caption{Our system framework. The system takes RGB images and depth images as input (A). First, information extraction is performed (B), followed by plane processing (C). After that, data association of multiple information is conducted (D). Camera pose optimization is then executed using a factor graph with multiple constraints (E). Finally, plane fusion and updating take place (F). The system outputs camera poses and a plane map (G).}
 	\label{framework}
 	\vspace{-5.5mm} % 控制整个图（包括标题）与下面的距离
 \end{figure*}

% 针对以上问题，我们提出了一套面向复杂环境的基于平面特征的视觉SLAM系统，包含平面信息处理，数据关联和多约束优化。首先，使用PEAC算法 提取平面，针对平面提取的不准确性和不稳定性问题，我们结合语义、平面结构等信息对平面进行处理；然后，我们集成了平面参数、语义、IoU以及非参数检验进行平面的数据关联，与常规方法相比，我们的方法在平面结构复杂的场景下具有准确性和稳健性的显著优势；最后，我们耦合了点-点、面-面、点-面、面-语义的关联约束以及平面的结构约束，构建了多约束因子图对相机位姿进行优化。我们使用公共室内数据集对我们的所提出的系统性能进行了评估，实验结果表明，与最先进的相关研究方法相比，我们的系统在复杂场景下的建图效果（如图1（d）所示）和相机定位精度都达到了具有竞争力的水平。
To address the aforementioned issues, we propose a visual SLAM system based on planar features tailored for planar ambiguous scenes, encompassing plane information processing, data association, and multi-constraint optimization. The system framework is shown in Fig.\ref{framework}. Initially, we employ the PEAC\cite{feng2014fast} algorithm for plane extraction. To address the issues related to inaccurate and unstable plane extraction, we combine semantic, planar structure, and other information to process the plane. Subsequently, we integrate plane parameters, semantics, IoU, and non-parametric tests to establish associations between planes. Compared to conventional methods, our approach has a significant advantage in accuracy and robustness in planar ambiguous scenes. Finally, we couple plane-to-plane, point-to-plane, plane-to-semantic association constraints, along with structural constraints for planes, to construct a multi-constraint factor graph for optimizing camera poses. We evaluated the performance of our proposed system using publicly available indoor datasets, and the experimental results demonstrate that our system achieves competitive levels of mapping (as shown in Fig.\ref{Overview}(d)) and camera localization accuracy in planar ambiguous scenes when compared to state-of-the-art research methods.

% 我们的贡献总结如下：
% ① 我们首次将语义信息用于平面处理，并提取规则物体平面的边缘线和顶点，以应用于平面筛选、数据关联和位姿优化等环节；
% ② 我们提出了一种针对平面特征的集成数据关联策略，以保证在复杂场景下平面数据关联的准确性；
% ③ 得益于准确、稳定的平面筛选、数据关联和多约束优化，我们的系统在复杂场景下的地图构建和相机定位精度得到提升。
% ④ 据我们所知，我们是第一个在复杂场景下进行精细化平面地图构建的工作。
\textbf{The main contributions of this paper are as follows:}
\begin{itemize}
	\item We incorporate semantic information into plane processing and extract edge lines and vertices of planes from regularly shaped objects, which are applied in plane selection, data association, and pose optimization. 
	\item We propose an integrated data association strategy for planar features to ensure the accuracy of planar data association in planar ambiguous scenes.
	\item Due to precise and stable plane processing, data association, and multi-constraint optimization, our system enhances the accuracy of map construction and camera localization in planar ambiguous scenes.
 	% \item To the best of our knowledge, we are the first to work on building detailed planar maps under planar ambiguous scenes.
\end{itemize}

% ---->>>>>>>>>>>>> 图1 系统概述
 
%%%%%%%%%%%%%%%%%%%%%%%%%%%%%%%%%%%%%%%%%%%%%%%%%%%%%%%%%%%%%%%%%%%%%%%%%%%%%%%%
\section{Related Work}
%%%%%%%%%%%%%%%%%%%%%%%%
\subsection{Visual SLAM Based on Planar Features}
% 平面特征在结构化环境中大量存在，对于位姿约束和建图都具有重要作用。CPA-SLAM 通过直接图像对准关键帧和EM框架中的全局平面模型来一致跟踪相机运动。SP-SLAM 并利用平面边缘约束生成假定平面，同时在优化中增加平面的垂直约束和平行约束；Huang等人 从后端的关键帧中提取平面，合并过度分割的平面，在g2o中优化关键帧和全局地标平面的姿态。Li等人 提出了一种新的基于RGB图像的两阶段平面检测策略和一种新的共面点和线的参数化方法；TT-SLAM 利用基于平面模板的跟踪器（TT）来计算相机姿态并重建多平面场景表示.Shu等人 在基于特征的平面SLAM系统中引入了一种基于能量的几何模型拟合方法，将SLAM视为优化不同类型的几何多模型估计；VIP-SLAM 引入平面信息来减少地图点的数量，加快BA的优化，并将平面信息集成到整个SLAM系统中，实现高精度跟踪。Structure PLP-SLAM 使用点和线进行稳健的相机定位，并同时对环境进行平面重建，以实时提供结构图；
\textbf{Planar features exist abundantly in structured environments and play an important role in pose constraints and mapping.} SP-SLAM\cite{zhang2019point} utilizes planar edge constraints to hypothesize planes while adding orthogonal and parallel constraints on planes during optimization. Huang et al.\cite{huang2019optimization} extract planes from backend keyframes and optimize keyframe and landmark plane poses in g2o\cite{kummerle2011g}. Li et al.\cite{li2020co} propose a new parametrization method for coplanar points and lines. Shu et al.\cite{shu2021visual} introduce an energy-based geometric model fitting method in feature-based planar SLAM. VIP-SLAM\cite{chen2022vip} incorporates planar information to reduce map points, speed up BA optimization. Structure PLP-SLAM\cite{shu2023structure} uses points and lines for robust camera localization while concurrently reconstructing planar environments.

% 还有些工作在平面SLAM中进一步利用曼哈顿假设来提供额外结构约束。Kim 等人  基于曼哈顿世界假设求解相机的旋转部分，并采用卡尔曼滤波估计相机位姿以及更新地图平面。Structure-SLAM 和PlanarSLAM 采用基于均值漂移聚类算法提取曼哈顿坐标系，解耦旋转和平移，利用MW假设进行旋转估计；ManhattanSLAM 采用正交平面法提取曼哈顿坐标系，检查平面之间的正交关系以直接检测曼哈顿帧，将场景建模为曼哈顿帧的混合；
\textbf{Some other work further utilizes the MW (Manhattan world) assumption to provide additional structural constraints in planar SLAM.} Kim et al.\cite{kim2018linear,joo2021linear} solve the rotational part of the camera pose based on the MW assumption and use Kalman filtering to estimate the camera pose and update map planes. Structure-SLAM\cite{li2020structure} and PlanarSLAM\cite{li2021rgb} extract the Manhattan coordinate system using mean shift clustering and decouple rotation and translation. ManhattanSLAM\cite{yunus2021manhattanslam} checks orthogonal relationships between planes to directly detect Manhattan frames, and models the scene as a mixture of Manhattan frames.

% 这些工作所针对的场景平面结构比较简单，未考虑在实际环境中存在平面混淆的情况。
The scenes typically targeted by these works usually have relatively simple planar structures. They do not consider the presence of ambiguous planes in real-world environments.

\subsection{Data Association in Planar SLAM}
% 数据关联是平面SLAM不可或缺的组成部分，用于确定地图中已经构建的平面路标与当前帧检测到的平面之间的对应关系。CPA-SLAMxv直接比较平面参数进行平面的关联；Zhang等人 采用最近邻搜索方法来获得潜在的平面对应关系，对每个潜在平面对应关系进行进一步的重叠检查。Pop-up slam 基于平面参数以及平面之间的投影重叠进行平面关联；Hsiao等人 使用平面参数结合地标平面模型拟合测量平面的残差进行平面关联；Yang等人 首先比较了平面的参数信息，然后将具有最多共享特征点的平面进行匹配；Zhang等人xvi针对平面较小的角度误差容易引起较大与坐标系原点的距离误差的问题，首先比较平面法向量的角度，然后根据一个平面的边缘点到其他平面的距离进行平面的关联；在后续的研究工作 中，根据相交的直线计算平面后，计算从直线的端点到平面地标的平均距离，与平面法向量结合进行平面匹配。PlanarSLAM和ManhattanSLAM沿用了Zhang等人的数据关联策略。然而，在具有复杂平面结构的场景中，平面之间的参数信息相差很小，基于投影重叠、共享特征点、边缘点到平面的距离等策略也很难起到作用，这些方法很难做到平面特征的准确关联。
Data association is an indispensable part of planar SLAM, used to determine the correspondence between planes constructed in the map and planes detected in the current frame. CPA-SLAM\cite{ma2016cpa} directly compares plane parameters for plane association. Zhang et al.\cite{zhang2016point} use nearest neighbor search and overlap check for plane association. Pop-up SLAM\cite{yang2016pop} associates planes based on plane parameters and projection overlap between planes. Hsiao et al.\cite{hsiao2017keyframe} combine plane parameters with fitting residuals of measuring planes for plane association. Yang et al.\cite{yang2019monocular} first compare plane parameter, then match planes with the most shared feature points. Zhang et al.\cite{zhang2019point} first compare the plane normals, and then associate the planes based on the distances from the edge points of one plane to other planes. In subsequent work\cite{zhang2021stereo}, they combine the average distance from the line endpoints to the landmark planes with the plane normals for plane matching. PlanarSLAM\cite{li2021rgb} and ManhattanSLAM\cite{yunus2021manhattanslam} adopt the data association strategy from \cite{zhang2019point}. However, in planar ambiguous scenes, the parameter differences between planes are very small. Strategies based on projection overlap, shared feature points, edge point to plane distances, etc. also struggle to work. These methods have difficulty achieving accurate association of planes.

\section{Specific Implementation} 
%%%%%%%%%%%%%%%%%%%%%%%%
\subsection{Information Extraction}
% 此部分对应图2A模块。我们的系统基于ORB-SLAM2算法构建，仍使用ORB特征进行点特征提取。对于平面特征，基于PEAC算法进行提取，得到平面的点云和在当前帧下的参数。为节省计算量并增加对平面信息的感知，提取平面的边缘点。在复杂场景下，场景中存在大量的物体，可以在有些形状规则的物体上提取到平面；但平面检测的准确性并不稳定，容易出现欠分割和过分割；此外规则物体上提取到的平面，参数信息与放置该物体的平面十分相似，为平面的数据关联带来了严峻的挑战。针对以上问题，我们基于YOLOX目标检测算法提取物体的语义信息，并基于多目标跟踪算法进行目标检测的漏检补偿，服务于平面筛选、数据关联等。
This section corresponds to Fig.\ref{framework}.(B). Our system is built upon the ORB-SLAM2\cite{mur2017orb} algorithm, and still utilizes ORB features for point feature extraction. For planar features, we extract them based on the PEAC\cite{feng2014fast} algorithm, obtaining point clouds of the planes and their parameters. To reduce computational load and enhance the perception of plane information, edge points of the planes are extracted. In planar ambiguous scenes, where numerous objects are present, planes can be extracted from certain geometrically regular objects. However, the accuracy of plane detection is not stable and prone to under- and over-segmentation. Additionally, the parameters of planes extracted from regularly shaped objects are highly similar to the planes where these objects are placed, posing a serious challenge to the correlation of planes. To address the aforementioned issues, we extract semantic information of objects based on the YOLOX\cite{ge2021yolox} to serve plane screening and data correlation.

\subsection{Plane Processing}
% 此部分对应图2.C，具体由以下五个模块组成：
This section corresponds to Fig.\ref{framework}.(C) and is specifically composed of the following five modules:
% 平面二次拟合：基于PEAC算法提取到平面信息后，使用PCL点云库对平面进行二次拟合，以得到更准确的平面参数和点云，然后对平面的边缘点根据点到平面的距离进行筛选。
\subsubsection{\bf{Plane Re-fitting}}
After extracting plane information based on the PEAC algorithm, we employ the PCL (point cloud library) to perform a re-fitting of the planes. This step aims to obtain more precise plane parameters and point clouds. Subsequently, we filter the edge points of the planes.

% 平面与目标检测框关联：将筛选后的平面边缘点投影到图像平面上，构建投影矩形框。然后依据平面投影框与目标检测框的IoU以及边缘点在目标检测框中的比例，构建平面与目标检测框之间的关联，为平面赋予物体类别ID，没有关联物体的平面类别ID设置为-1。
\subsubsection{\bf{Associating Planes with Object Detection Boxes}}
After fitting the planar edge points, we project them onto the image plane to construct a projection rectangle box. Then, based on the IoU between the projection box and the object detection box as well as the proportion of edge points in the object detection box, we establish the association between the plane and the object detection box, assigning the object category ID to the plane. For planes without associated objects, the category ID is set to -1.

% 平面边缘线提取：存在目标检测框关联的平面一般都是键盘、书本、显示器等形状规则的物体上检测出来的平面，对于这些平面，我们基于随机样本一致性(RANSAC)从平面的边缘点中提取多条直线：
% 其中，$X_{\text{inliers}}$表示直线内点集,$l(\theta)$表示参数化直线模型,$d(x_i, l(\theta))$表示点$x_i$到直线$l$的距离。通过迭代求解,可获得使内点数最大的直线模型。直线参数化为6维向量$L={{({{P}^{T}},{{V}^{T}})}^{T}}$,表示线上的一点坐标$P={{({{P}_{x}},{{P}_{y}},{{P}_{z}})}^{T}}$和归一化的方向向量$V={{({{V}_{x}},{{V}_{y}},{{V}_{z}})}^{T}}$。
\subsubsection{\bf{Extracting Edge Lines of Planes}}
Planes that are associated with object detection boxes are generally detected on regularly shaped objects. For these planes, we extract multiple lines from the edge points of the planes using the Random Sample Consensus (RANSAC) algorithm.
% \begin{equation}
% % loss(\theta) = \sum_{x_i\in X_{\text{inliers}}} |d(x_i, l(\theta))|^2,
% loss(\theta )=\sum\nolimits_{{{x}_{i}}\in {{X}_{\text{inliers}}}}{{{\left| d({{x}_{i}},l(\theta )) \right|}^{2}}},
%     \label{eq1}
% \end{equation}
% where $X_{\text{inliers}}$ represents the set of inlier points on the line, $l(\theta)$ represents the parameterized line model, and $d(x_i, l(\theta))$ represents the distance from point $x_i$ to the line $l$. By iteratively solving this equation, we can obtain the line model that maximizes the number of inlier points. 
The parameterization of the line is represented as a 6-dimensional vector $L={{({{P}^{T}},{{V}^{T}})}^{T}}$, where it includes a point on the line $P$ and a normalized direction vector $V$.

% 平面顶点提取：然后基于方向向量判断直线之间的平行关系，并进一步提取平面的4个顶点。（规则物体的平面的4个顶点是平面边缘提取到的4条直线的交点，但是平面和平面边缘点提取的质量无法保证，且直线的提取存在误差，直接联立直线方程求交点可能会因为直线提取的误差导致偏差，甚至无法求解。所以，）利用两空间直线的最小公垂线的垂足间接求得两直线的交点，这种方法对误差具有更强的鲁棒性。对于两条不平行的直线${{L}_{i}}$和${{L}_{j}}$，假设公垂线与直线${{L}_{j}}$和直线${{L}_{i}}$的交点分别为：
\subsubsection{\bf{Extraction Vertices of Planes}}
Using directional vectors to detect line parallelism and extract the four vertices of a plane.
% For regular objects, the four vertices of the plane are the intersections of the four lines extracted from the plane edges. However, the quality of plane extraction and edge point extraction cannot be guaranteed, and there may be errors in line extraction. Directly solving for the intersection points using the equations of the lines may result in deviations, or even be unsolvable due to errors in line extraction. Therefore, we 
We indirectly calculate the intersection points of two lines using the minimum common perpendicular line between them, which provides greater robustness against errors. For two non-parallel lines ${L}_{i}$ and ${L}_{j}$, assuming that the intersection points of the common perpendicular line with lines ${L}_{j}$ and ${L}_{i}$ are:
\begin{equation}
{{P}_{pj}}={{P}_{j}}+{{K}_{j}}*{{V}_{j}},
    \label{eq2}
\end{equation}
\begin{equation}
{{P}_{pi}}={{P}_{i}}+[({{P}_{pj}}-{{P}_{i}})\cdot {{P}_{i}}]*{{V}_{i}},
    \label{eq3}
\end{equation}
% 其中，\[{{K}_{j}}\]为系数，代表公垂线与直线${{L}_{j}}$的交点与P_j的距离比例。根据向量\[{{P}_{pi}}-{{P}_{pj}}\]与${{V}_{j}}$垂直：
where ${{K}_{j}}$ is a coefficient representing the ratio of the intersection point of the common perpendicular line with line ${{L}_{j}}$ to ${{P}_{j}}$. According to the orthogonality between the vector ${{P}_{pi}}-{{P}_{pj}}$ and ${{V}_{j}}$, we can calculate:
\begin{equation}
{{K}_{j}}=\frac{\left( {{P}_{i}}-{{P}_{j}}+(({{P}_{j}}-{{P}_{i}})\cdot {{V}_{i}})*{{V}_{i}} \right)\cdot {{V}_{j}}}{{{V}_{j}}\cdot {{V}_{j}}-({{V}_{i}}\cdot {{V}_{j}})*({{V}_{j}}\cdot {{V}_{i}})}.
    \label{eq5}
\end{equation}

% 由此可以求得两直线${{L}_{i}}$和${{L}_{j}}$共垂线的两个垂足点\[{{P}_{pi}}\]和\[{{P}_{pj}}\]。当两直线${{L}_{i}}$和${{L}_{j}}$共面时，两个垂足点\[{{P}_{pi}}\]和\[{{P}_{pj}}\]应当重合。当满足以下三个条件时，证明该平面顶点提取成功：①两个垂足点\[{{P}_{pi}}\]和\[{{P}_{pj}}\]之间的欧氏距离小于一定阈值；②\[{{P}_{pi}}\]和\[{{P}_{pj}}\]到该平面的距离在一定范围内；③\[{{P}_{pi}}\]和\[{{P}_{pj}}\]的投影点落在该平面关联的目标检测框内。
From this, we can obtain the two perpendicular foot points ${{P}_{pi}}$ and ${{P}_{pj}}$ of the common perpendicular line of ${{L}_{i}}$ and ${{L}_{j}}$. When ${{L}_{i}}$ and ${{L}_{j}}$ are coplanar, ${{P}_{pi}}$ and ${{P}_{pj}}$ should coincide. The successful extraction of the plane vertices is verified when the following three conditions are met: (1) The Euclidean distance between ${{P}_{pi}}$ and ${{P}_{pj}}$ is less than a certain threshold. (2) The distance from ${{P}_{pi}}$ and ${{P}_{pj}}$ to the associated plane is within a certain range. (3) The projection points of ${{P}_{pi}}$ and ${{P}_{pj}}$ fall within the object detection box associated with the plane.

\subsubsection{\bf{Plane Selection}}
% 平面筛选：准确的平面信息为后续数据关联、多约束优化提供了稳定可靠的输入，是保证位姿估计和地图构建的准确性的基础。但是，平面检测算法的准确性和稳定性难以保证，所以我们结合深度、语义、结构、内点比例等多种信息对平面进行筛选，将以下平面设置为bad：①距离当前观测位置比较远；②通过PCL点云库二次拟合得到的内点占比较小；③与规则物体的目标检测框构建了关联，但该平面在图像边缘附近，或者没有提取到2对平行直线或4个顶点；④边缘点落在人、泰迪熊等目标检测框内的比例较大。
Accurate plane information provides stable and reliable input for subsequent data correlation and multi-constraint optimization, and is the basis for ensuring the accuracy of pose estimation and map construction. However, the accuracy and stability of plane detection algorithms are difficult to guarantee. Therefore, we combine multiple information such as depth, semantics, structure, and inner point proportion to screen planes, and categorize the following plane as ``bad": (1) Planes that are relatively far away from the current observation position; (2) Planes with a small proportion of inner points obtained through re-fitting using the PCL; (3) Planes that are associated with the bounding box of regularly shaped objects, but are located near the image edges or no parallel pairs of lines or four vertices are extracted; (4) Planes where a large proportion of edge points fall within the bounding box of unstructured objects.

%%%%%%%%%%%%%%%%%%%%%%%%
\subsection{Data Association}
% 我们系统的数据关联主要涉及点-点、点-面、面-面、目标检测框与平面投影框，如图2.D所示。其中，地图点与特征点的关联沿用了ORB-SLAM2的策略；地图点与frame-plane的关联主要基于点到平面的距离，并使用投影点与目标检测框的位置关系以及平面与目标检测框的关联关系进行辅助关联；目标检测框与map-plane投影框的关联只针对属于形状规则物体的map-plane，在关联之前，将该map-plane的顶点或者边缘点投影到图像上构建该平面的投影框，然后基于IoU构建与目标检测框的关联。
Our system's data association primarily involves point-plane, object detection box-plane projection box, and plane-plane associations, as illustrated in Fig.\ref{framework}.(D). 

% \subsubsection{\bf{Point-Point}}
% For the association between map points and feature points, we adopt the ORB-SLAM2 strategy.

\subsubsection{\bf{Point-Plane}}
For the association between map points and frame planes, it is mainly based on the point-to-plane distance. We also use the position relationship between the projection points and object detection boxes, as well as the association relationship between the planes and object detection boxes, to assist in the association.

\subsubsection{\bf{Object Detection Box-Plane Projection Box}}
The association between object detection boxes and map-plane projection boxes is only applicable to map planes with a category ID not equal to -1. Before establishing the association, we project the vertices or edge points of this map plane onto the image to construct the projection box for that plane. Then, we establish the association based on the IoU between the projection box and the detection box.

\subsubsection{\bf{Plane-Plane}}
% 对检测到的平面进行筛选后，首先进行平面地图初始化，平面地图初始化完成后，对frame-plane以及map-plane进行数据关联。将平面参数化表达为海塞形式$\pi ={{({{n}^{T}},d)}^{T}}$，其中$n={{({{n}_{x}},{{n}_{y}},{{n}_{z}})}^{T}}$表示平面的单位法向量，$d$表示平面到坐标系原点的距离。记当前帧检测到的某个平面实例为${{P}_{c}}$，参数为${{\pi }_{c}}={{({{n}_{c}}^{T},{{d}_{c}})}^{T}}$，地图中的某个平面实例为${{P}_{w}}$，参数为${{\pi }_{w}}={{({{n}_{w}}^{T},{{d}_{w}})}^{T}}$。如果检测平面$P_{i}^{D}$与地图平面$P_{j}^{M}$是同一个平面，其法向量的方向应当是非常接近的，基于这个事实，首先判断平面${{P}_{c}}$和${{P}_{w}}$的法向量之间的角度$\beta $：
% $\beta =\arccos (\frac{\left| {{({{R}_{cw}}^{-1}{{n}_{c}})}^{T}}\cdot {{n}_{w}} \right|}{\left| {{R}_{cw}}^{-1}{{n}_{c}} \right|\left| {{n}_{w}} \right|})<{{\beta }_{Th}}$
% 	其中Tcw∈R4*4代表初步估计的世界坐标系到当前帧相机坐标系之间的变换矩阵，包含旋转矩阵Rcw∈R3*3和平移向量Rcw∈R3*1，${{\beta }_{Th}}$表示设定的角度阈值。经过法向量的角度筛选后，进行平面${{P}_{c}}$和${{P}_{w}}$之间的距离d的判断：
% $d=\left| {{t}_{cw}}^{T}{{n}_{c}}+{{d}_{c}}-{{d}_{w}} \right|<{{d}_{Th}}$
% 其中${{d}_{Th}}$表示设定的平面之间的距离参数。为避免平面角度误差使得d误差过大导致关联失败，进一步比较平面${{P}_{c}}$的边缘点${{B}_{c}}$到平面${{P}_{w}}$的距离：
% $R=\frac{\left\| ({{n}_{w}}^{T}({{R}_{cw}}^{-1}B_{c}^{i}-{{R}_{cw}}^{-1}{{t}_{cw}})+{{d}_{w}}<{{d}_{Th}}^{\prime })_{i=1}^{M} \right\|}{M}>{{R}_{Th}}$
% 其中M表示平面${{P}_{c}}$的边缘点${{B}_{c}}$的个数，${{d}_{Th}}^{\prime }$表示设定的点到平面的距离阈值，$\left\| \centerdot  \right\|$表示满足条件的点的个数，$R$表示满足条件的边缘点比例，${{R}_{Th}}$表示设定的比例阈值。
We represent plane parameters as Hessian form $\pi =({n^{T}},d)^{T}$, where $n=({n_{x}},{n_{y}},{n_{z}})^{T}$ represents the unit normal vector of the plane, and $d$ represents the distance from the plane to the origin of the coordinate system. Denote a frame-plane as $P_{c}$ with parameters $\pi_{c}=({n_{c}}^{T},{d_{c}})^{T}$, and a map-plane as $P_{w}$ with parameters $\pi_{w}=({n_{w}}^{T},{d_{w}})^{T}$. 
% If $P_{c}$ and $P_{w}$ are the same, their normal vectors should be very close. Based on this fact, 
We first calculate the angle $\beta$ between the normals of plane ${{P}_{c}}$ and ${{P}_{w}}$:
\begin{equation}
% \arccos\left(\frac{\left| {{({R}_{cw}^{-1}{{n}_{c}})}^{T}}\cdot {{n}_{w}} \right|}{\left| {R}_{cw}^{-1}{{n}_{c}} \right|\left| {{n}_{w}} \right|}\right) < {{\beta }_{Th}},
\arccos ({\left| {{({{R}_{cw}}^{-1}{{n}_{c}})}^{T}}\cdot {{n}_{w}} \right|}/{\left| {{R}_{cw}}^{-1}{{n}_{c}} \right|\left| {{n}_{w}} \right|}\;)<{{\beta }_{T}},
    \label{eq6}
\end{equation}
% $R_{cw} \in \mathbb{R}^{3 \times 3}$ represents the rotation matrix between the world coordinate system and the camera coordinate system.
% where $T_{cw} \in \mathbb{R}^{4 \times 4}$ represents the estimated transformation matrix from the world coordinate system to the camera coordinate system for the current frame. It includes the rotation matrix $R_{cw} \in \mathbb{R}^{3 \times 3}$ and the translation vector $t_{cw} \in \mathbb{R}^{3 \times 1}$.
where $R_{cw} \in \mathbb{R}^{3 \times 3}$ represents the rotation matrix between the world coordinate system and the camera coordinate system. ${{\beta }_{T}}$ denotes the specified angle threshold. After the angle filtering based on the normals, we proceed to evaluate the distance $d$ from the origin of the world coordinate system to plane ${{P}_{c}}$ and ${{P}_{w}}$:
\begin{equation}
\left| {t}_{cw}^{T}{{n}_{c}} + {{d}_{c}} - {{d}_{w}} \right| < {{d}_{T}},
    \label{eq7}
\end{equation}
where $t_{cw} \in \mathbb{R}^{3 \times 1}$ represents the translation vector, ${{d}_{T}}$ represents the difference threshold in distance from the origin of the world coordinate system to two planes. To avoid large errors in $d$ due to plane angle discrepancies that might cause association failure, taking reference from \cite{zhang2019point}, we further compare the distances between the edge points $B_{c}$ of plane ${{P}_{c}}$ and plane ${{P}_{w}}$:
\begin{equation}
% \frac{\left\| ({n}_{w}^{T}({R}_{cw}^{-1}B_{c}^{i} - {R}_{cw}^{-1}{{t}_{cw}}) + {{d}_{w}} < {d}_{Th}')_{i=1}^{M} \right\|}{M} > {{R}_{Th}},
\left\| (n_{w}^{T}(R_{cw}^{-1}B_{c}^{i}-R_{cw}^{-1}{{t}_{cw}})+{{d}_{w}}<{{d}_{T}'})_{i=1}^{m} \right\|/m\ >{{R}_{T}},
    \label{eq8}
\end{equation}
where $m$ represents the number of edge points $B_{c}$ of plane ${{P}_{c}}$, ${d}_{T}'$ denotes the specified point-to-plane distance threshold, $\left\| \centerdot  \right\|$ represents the count of points that satisfy the condition, and ${{R}_{Th}}$ represents the specified ratio threshold. 

% 当平面${{P}_{c}}$和${{P}_{w}}$的类别ID均为-1时，如果满足角度筛选条件（1），同时满足条件（2）或者（3），则判定平面${{P}_{c}}$和${{P}_{w}}$关联成功。当平面${{P}_{c}}$和${{P}_{w}}$的类别ID均不为-1时，在满足满足角度筛选条件（1）的情况下，计算${{P}_{w}}$在图像上的投影框与${{P}_{c}}$关联的目标检测框的IoU，以关联${{P}_{c}}$和${{P}_{w}}$。但是在某些相机视角下由于物体重叠会导致该方法关联错误，为此，使用非参数检验进行辅助关联。
% 将平面${{P}_{c}}$和${{P}_{w}}$关联的地图点集合分别记为${{M}_{c}}$和${{M}_{w}}$，${{M}_{c}}$和${{M}_{w}}$的数量分别为$I$和$J$，构建Mann-Whitney统计量：
% $W=min(\sum\limits_{i=1}^{I}{R_{i}^{c}}-\frac{I(I+1)}{2},\sum\limits_{j=1}^{J}{R_{j}^{w}}-\frac{J(J+1)}{2})$
% 其中$R_{i}^{c}$、$R_{j}^{w}$分别为${{M}_{c}}$和${{M}_{w}}$的坐标值在混合样本集合${{M}_{c}}\cup {{M}_{w}}$中的秩。通过大样本近似可以证明统计量$W$的渐近分布为正态分布，其均值和方差为：
% $E(W)=I(I+J+1)/2$
% $D(W)=\frac{IJ(I+J+1)}{12}-\frac{IJ(\sum\limits_{k=1}^{g}{\tau _{k}^{3}}-\sum\limits_{k=1}^{g}{{{\tau }_{k}}})}{12(I+J)(I+J-1)}$
% 其中，${{\tau }_{k}}\in {{M}_{c}}\cap {{M}_{w}}$代表${{M}_{c}}$和${{M}_{w}}$中的重复值。为了使得${{M}_{c}}$和${{M}_{w}}$分布相同的假设成立，在拒绝域选取为$\alpha$时，即置信区间的临界值为${{z}_{\alpha /2}}$时，统计量$W$应满足：
% $S_{ij}^{np}=[\frac{\left| E(W)-W \right|}{\sqrt{D(W)}}<{{z}_{\alpha /2}}]$
% 当统计量$W$满足式*时，可以说明${{M}_{c}}$和${{M}_{w}}$分布相同，${{P}_{c}}$和${{P}_{w}}$关联成功。需要注意的是，当$I$和$J$较小时，非参数检验可靠性较低无法使用。当${{P}_{c}}$和${{P}_{w}}$通过IoU关联上且非参数检验可以使用时，使用非参数检验对关联结果进行验证。
When the class IDs of ${{P}_{c}}$ and ${{P}_{w}}$ are both -1, the association between ${{P}_{c}}$ and ${{P}_{w}}$ is considered successful if it satisfies $\text{Eq.(\ref{eq6})} \cap (\text{Eq.(\ref{eq7})} \cup \text{Eq.(\ref{eq8})})$. When the class IDs of ${{P}_{c}}$ and ${{P}_{w}}$ are both not -1, under the condition that the angle filtering condition Eq.(\ref{eq6}) is satisfied, the IoU between the projection box of ${{P}_{w}}$ and the object detection box associated with ${{P}_{c}}$ is calculated to associate ${{P}_{c}}$ and ${{P}_{w}}$. However, under certain camera perspectives, overlapping objects can cause association errors by this method. 

% 为此，受EAO-SLAM的物体级数据关联启发，我们将非参数检验用于平面的辅助关联。
% 收到 EAO-SLAM 的物体级数据关联启发，我们将其应用在平面关联上
% For this purpose, inspired by EAO-SLAM, we apply non-parametric tests for the auxiliary association of planes. 
For this purpose, inspired by the object-level data association in EAO-SLAM\cite{wu2020eao}, we employ non-parametric tests to assist in plane association.
% For this purpose, non parametric tests are used for auxiliary correlation. 
The sets of map points associated with ${{P}_{c}}$ and ${{P}_{w}}$ are denoted as ${{M}_{c}}$ and ${{M}_{w}}$ respectively. The quantities of ${{M}_{c}}$ and ${{M}_{w}}$ are $I$ and $J$ respectively. A Mann-Whitney statistic is constructed as follows:
\begin{equation}
W = \min \left( \sum\limits_{i=1}^{I}{R_{i}^{c}} - \frac{I(I+1)}{2}, \sum\limits_{j=1}^{J}{R_{j}^{w}} - \frac{J(J+1)}{2} \right),
    \label{eq9}
\end{equation}
where ${{R_{i}^{c}}}$ and ${{R_{j}^{w}}}$ represent the ranks of the coordinate values of ${{M}_{c}}$ and ${{M}_{w}}$ within the mixed sample set ${{M}_{c}} \cup {{M}_{w}}$. It can be shown that the asymptotic distribution of the statistic $W$ is normal\cite{wilcoxon1992individual}, with a mean and variance of:
\begin{equation}
E(W) = {I(I+J+1)}/{2},
    \label{eq10}
\end{equation}
\begin{equation}
D(W) = \frac{IJ(I+J+1)}{12} - \frac{IJ\left(\sum\limits_{k=1}^{g}{\tau _{k}^{3}} - \sum\limits_{k=1}^{g}{{{\tau }_{k}}} \right)}{12(I+J)(I+J-1)},
    \label{eq11}
\end{equation}
where ${{\tau }_{k}}\in {{M}_{c}}\cap {{M}_{w}}$ represents duplicate values in ${{M}_{c}}$ and ${{M}_{w}}$. To uphold the assumption that ${{M}_{c}}$ and ${{M}_{w}}$ have the same distribution, with the critical value for the confidence interval to be ${{z}_{\alpha /2}}$, the statistic $W$ should satisfy:
\begin{equation}
S_{ij}^{np} = \left[ \frac{\left| E(W)-W \right|}{\sqrt{D(W)}} < {{z}_{\alpha /2}} \right].
    \label{eq12}
\end{equation}

When the statistic $W$ satisfies Eq.(\ref{eq12}), it indicates that ${{M}_{c}}$ and ${{M}_{w}}$ have the same distribution, confirming the successful association between ${{P}_{c}}$ and ${{P}_{w}}$. Non-parametric tests may have lower reliability when $I$ and $J$ are small. When ${{P}_{c}}$ and ${{P}_{w}}$ are associated via IoU and non-parametric tests are applicable, perform non-parametric tests to validate the association results.

% \vspace{1mm}
%%%%%%%%%%%%%%%%%%%%%%%%
\subsection{Pose Optimization}
We propose a multi-constraint optimization factor graph to optimize camera poses, as illustrated in Fig.\ref{framework}.(E). The overall optimization function can be expressed as follows:
\begin{align}
{{T}_{cw}} = & \underset{{{T}_{cw}}}{\text{argmin}}\, \biggl( \sum{{H_{p}(f_{p})} + H_{\pi }(f_{\pi })} + H_{b\pi }(f_{b\pi }) \notag \\
&  + H_{p\pi }(f_{p\pi }) + H_{\pi p}(f_{\pi p}) + H_{\pi v}(f_{\pi v}) \biggr),
\label{eq13}
\end{align}
where $H(x)$ represents the Huber robust cost function, and $f$ denotes the error factors. The designed factor graph can be solved using the existing nonlinear optimization library g2o\cite{kummerle2011g}. The Eq.(\ref{eq13}) consists of the following components:

\subsubsection{\bf{Pose-Point Projection Factor}}
\begin{equation}
{{f}_{p}}({{p}_{w}},{{T}_{cw}})=\left\| {{u}_{obs}}-\rho ({{T}_{cw}}{{p}_{w}}) \right\|_{{{\sum }_{p}}}^{2},
    \label{eq14}
\end{equation}
where ${{u}_{obs}}$ represents the pixels associated with the map point ${{p}_{w}}$, $\rho ({{T}_{cw}}{{p}_{w}})$ denotes the projection of the map point ${{p}_{w}}$ from the world coordinate system onto the image, and $\sum $ represents the corresponding covariance matrix.

\subsubsection{\bf{Pose-Plane Projection Factor}}
\begin{equation}
{{f}_{\pi }}({{\pi }_{w}},{{T}_{cw}})=\left\| q({{\pi }_{c}})-q(T_{cw}^{-T}{{\pi }_{w}}) \right\|_{{{\sum }_{\pi }}}^{2},
    \label{eq15}
\end{equation}
where ${{\pi }_{c}}$ and ${{\pi }_{w}}$ represent the Hessian form $\pi ={{({{n}^{T}},d)}^{T}}$ of the associated ${P_{c}}$ and ${P_{w}}$ respectively,  and $T_{cw}^{-T}{{\pi }_{w}}$ denotes the transformation of the parameters of ${P_{w}}$ from the world coordinate system to the current camera coordinate system. Because representing a plane with three degrees of freedom using a four-dimensional vector in Hessian form would introduce additional computational complexity during optimization. Therefore, we use the minimal parameterization of the plane in optimization\cite{zhang2019point}:
\begin{equation}
q(\pi )=(\phi =\arctan \frac{{{n}_{y}}}{{{n}_{y}}},\varphi =\arcsin {{n}_{z}},d),
    \label{eq16}
\end{equation}
where $\phi $ and $\varphi $ are the azimuth and elevation angles of the normal vector.

\subsubsection{\bf{Box-Plane Projection Factor}}
\begin{equation}
{{f}_{b\pi }}({{V}_{w}},{{T}_{cw}})=\left\| {{B}_{obs}}-\eta ({{V}_{w}},{{T}_{cw}}) \right\|_{{{\sum }_{b\pi }}}^{2},
    \label{eq17}
\end{equation}
% 其中，\[{{V}_{w}}\]表示${{P}_{w}}$的投影框的最小4点表示，即哪四个${{P}_{w}}$的顶点或边缘点经过变换和投影后可以得到该投影框；\[\eta ({{V}_{w}},{{T}_{cw}})\]将这四个点\[{{V}_{w}}\]投影到图像上构建矩形框；\[{{B}_{obs}}\]表示与${{P}_{w}}$的投影框关联的目标检测框。
where ${{V}_{w}}$ represents the minimum 4-point representation of the projection box of ${P_{w}}$, i.e., the four vertices or edge points of ${P_{w}}$ that, after transformation and projection, form the projection box. $\eta ({{V}_{w}},{{T}_{cw}})$ represents the projection of these four points ${{V}_{w}}$ onto the image to construct a rectangular box. ${{B}_{obs}}$ represents the object detection box associated with ${P_{w}}$'s projection box.

\subsubsection{\bf{Point-Plane Association Factor}}
\begin{equation}
{{f}_{p\pi }}({{p}_{w}},{{T}_{cw}})=\left\| \pi _{c}^{T}({{T}_{cw}}{{p}_{w}}) \right\|_{{{\sum }_{p\pi }}}^{2},
    \label{eq18}
\end{equation}
where ${{p}_{w}}$ represents the map point associated with the plane ${P_{c}}$, and $\pi _{c}^{T}({{T}_{cw}}{{p}_{w}})$ represents the distance from the map point ${{p}_{w}}$, transformed to the current frame's camera coordinate system, to the plane ${P_{c}}$.

\subsubsection{\bf{Plane-Plane Parallel Factor}}
\begin{equation}
{{f}_{\pi p}}({{\pi }_{w}},{{T}_{cw}})=\left\| {{q}_{n}}({{n}_{c}})-{{q}_{n}}({{R}_{cw}}{{n}_{w}}) \right\|_{{{\sum }_{\pi p}}}^{2},
    \label{eq19}
\end{equation}
where ${{q}_{n}}$ represents the azimuth and elevation angles of the normal ${{n}_{c}}$. This error term represents the parallel constraint between planes.

\subsubsection{\bf{Plane-Plane Vertical Factor}}
\begin{equation}
{{f}_{\pi v}}({{\pi }_{w}},{{T}_{cw}})=\left\| {{q}_{n}}({{R}_{\bot }}{{n}_{c}})-{{q}_{n}}({{R}_{cw}}{{n}_{w}}) \right\|_{{{\sum }_{\pi v}}}^{2},
    \label{eq20}
\end{equation}
where ${{R}_{\bot }}{{n}_{c}}$ represents rotating the normal ${{n}_{c}}$ to the same direction as the normal ${{n}_{w}}$. 

\subsection{Plane Fusion and Update}
% 为解决漏关联的问题，在局部建图线程，根据上述的平面数据关联以及平面之间的共视关系，对地图中的平面进行融合。对于关联或者融合成功的平面进行更新，首先更新平面之间的共视关系，然后对平面的点云、边缘点云进行融合，最后对融合后的点云使用PCL点云库进行平面拟合，以更新平面的参数；此外，对于类别ID不为-1的平面，还需要进行顶点和边缘线的更新。
To address the issue of missed associations, within the local mapping thread, the map's planes are fused based on the previously mentioned plane data associations and the mutual visibility relationships between the planes, as illustrated in Fig.\ref{framework}.(F). For planes that are successfully associated or fused, we perform the following steps: updating mutual visibility relationships, fusing plane point clouds and edge point clouds, and updating plane parameters. Additionally, for planes with a category ID not equal to -1, vertex and edge line updates are also carried out.

%%%%%%%%%%%%%%%%%%%%%%%%%%%%%%%%%%%%%%%%%%%%%%%%%%%%%%%%%%%%%%%%%%%%%%%%%%%%%%%%
\section{Experiments and Results} 
% 在本节中，我们使用TUM RGB-D公共数据集对我们提出的系统进行了评估。由于我们主要研究基于平面特征的视觉SLAM在复杂场景下的问题，很多先进的相关工作缺乏相应的数据和效果展示，且没有开源，所以我们仅与ORB-SLAM2、SP-SLAM、PlanarSLAM和ManhattanSLAM这4个开源的相关研究方案进行对比，对比指标包含建图的定性效果、相机定位精度与实时性。PlanarSLAM和ManhattanSLAM不包含回环检测和全部BA模块，为公平对比，我们禁用了ORB-SLAM2、SP-SLAM和我们算法的以上两个模块。所有的实验均在配置为i9-12900H CPU，3060GPU和16GB内存的笔记本上执行，运行的操作系统为Ubuntu20.04。
In this section, we evaluate our proposed system using the TUM RGB-D\cite{sturm2012benchmark} dataset. 
Since our main focus is on the challenges of visual SLAM based on plane features in planar ambiguous scenes, many state-of-the-art related works lack corresponding data and performance demonstrations, and are not open-source. Therefore, we conduct comparisons with only four open-source related research solutions: ORB-SLAM2\cite{mur2017orb}, SP-SLAM\cite{zhang2019point}, PlanarSLAM\cite{li2021rgb} and ManhattanSLAM\cite{yunus2021manhattanslam}. The evaluation metrics include qualitative mapping performance, camera localization accuracy, and real-time performance. PlanarSLAM and ManhattanSLAM do not include the loop detection and full bundle adjustment (BA) module. To ensure a fair comparison, we disable the loop detection and full BA module in ORB-SLAM2, SP-SLAM, and our algorithm as well. All experiments are conducted on a laptop with an i9-12900H CPU, 3060GPU, and 16GB of memory.

\vspace{-3mm}

  \begin{figure}[h]
	% \vspace{-0.5mm}
\centering
 	\includegraphics[scale=0.24]{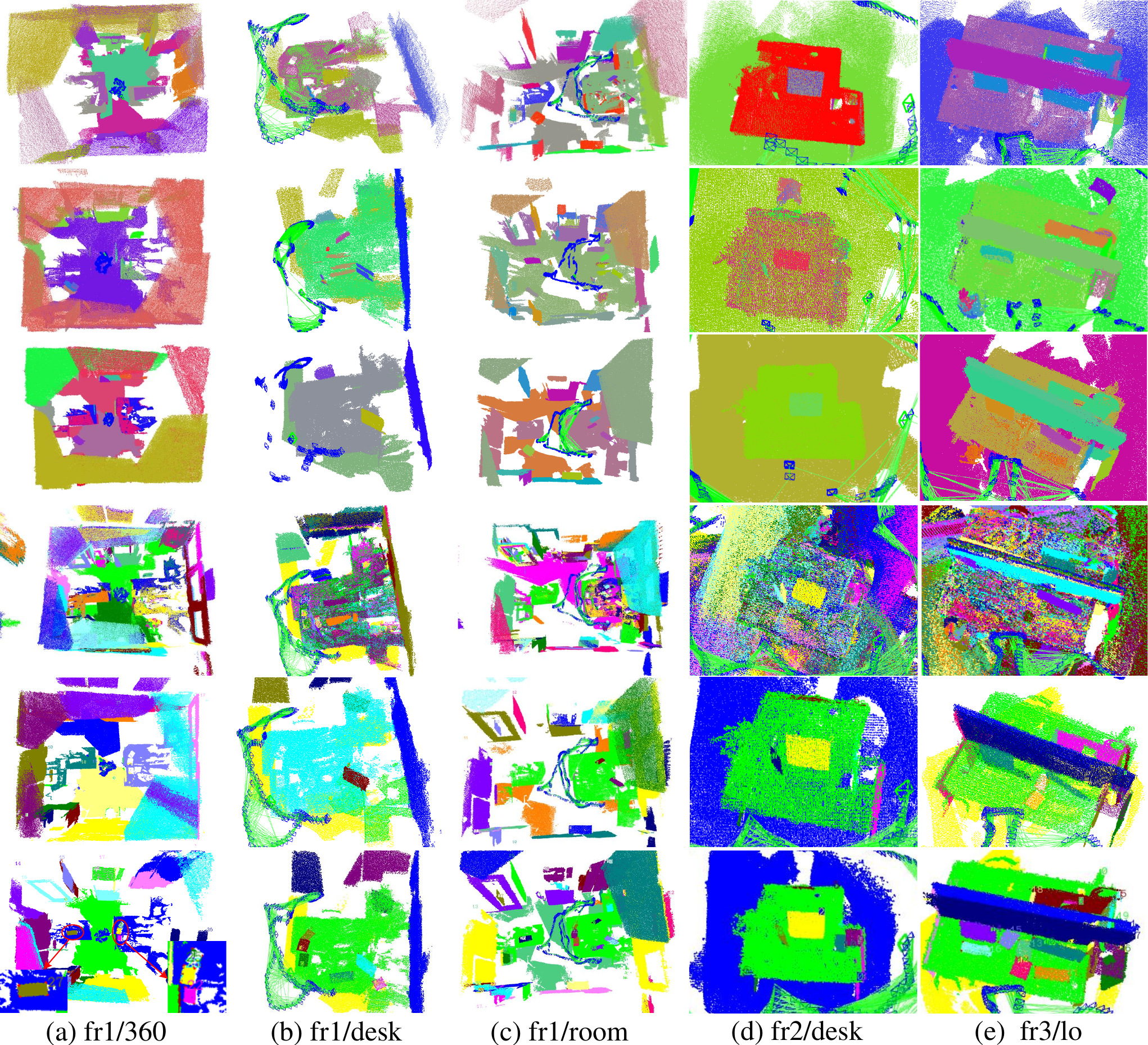}
 	\setlength{\abovecaptionskip}{-0.2cm} % 控制标题与图之间的距离
 	\caption{The plane mapping results for several algorithms. From top to bottom, they are SP-SLAM, PlanarSLAM, ManhattanSLAM,  w/o-(C)(D),  w/o-(D) and ours.}
 	\label{plane_map}
 	% \vspace{-2mm} % 控制整个图（包括标题）与下面的距离
 \end{figure}
 
\vspace{-3mm}

  \begin{figure}[h]
	\vspace{-0.5mm}
\centering
 	\includegraphics[scale=0.26]{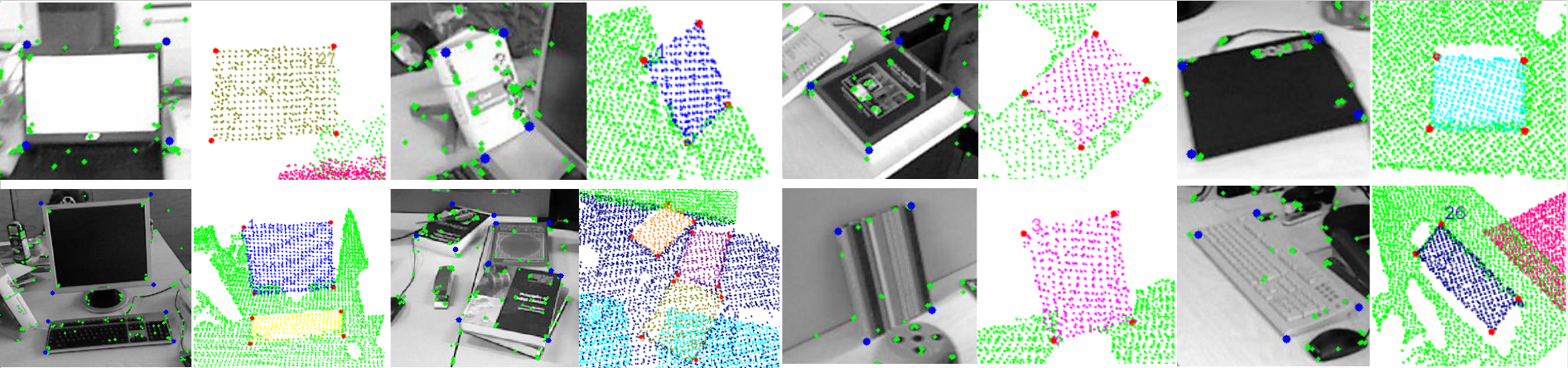}
 	\setlength{\abovecaptionskip}{-0.1cm} % 控制标题与图之间的距离
 	\caption{The results of plane vertex extraction. Each pair of images consists of a right-side image showing the vertices detected based on plane edge points and a left-side image displaying the projection of plane vertices onto the image.}
 	\label{Ver_Extract}
 	\vspace{-3mm} % 控制整个图（包括标题）与下面的距离
 \end{figure}
 
%%%%%%%%%%%%%%%%%%%%%%%%
\subsection{Mapping Qualitative Results}
% 我们的算法对平面顶点的提取效果如图*所示，展现了我们的算法对平面结构信息感知的准确性。
% 图*展示了在TUM数据集五个序列上SP-SLAM、PlanarSLAM、ManhattanSLAM与我们的算法的平面建图效果。其中， w/o-\ref{framework}.(C)(D)表示我们的算法不进行平面处理以及只使用参数对比的方式进行关联， w/o-\ref{framework}.(D)表示我们的算法只使用参数对比的方式进行关联，这两个实验对\ref{framework}.(C)和(D)进行了消融。在图*中，不同的平面使用不用的颜色表示。
% 对于其他序列，要么与上面这五种序列场景几乎相同，只是相机的运动方式不同，要么场景过于简单，所以没有给出对比效果展示。
% 可以看出SP-SLAM由于使用PCL拟合平面时选取的点云细粒度较低，因而对环境中的平面感知能力较弱，忽略了场景中的很多平面；PlanarSLAM和ManhattanSLAM与我们的算法均使用PEAC算法进行平面提取，选取的点云细粒度较高，可以提取环境中的更多平面，但是难以保证平面提取的质量，PlanarSLAM和ManhattanSLAM由于没有准确、鲁棒的平面筛选和关联策略，严重影响了地图构建的精度。例如PlanarSLAM在fr3-lo序列上构建的平面地图，没有将书本、键盘等物体上提取到的平面构建出来，反而在泰迪熊上构建出了很多小平面，这进一步影响了定位精度。
% 当我们的方法不进行平面处理以及只使用参数对比的方式进行关联时，由于平面提取和数据关联的质量下降，构建的平面地图非常杂乱；当进行平面处理但只使用参数对比的方式进行关联时，将物体和桌面提取到的平面进行了错误关联，导致无法实现精确的平面建图。
% 而我们的方法得益于准确的平面处理和关联策略，在这些复杂场景下构建了准确的平面地图。
 % The comparison of our algorithm's plane mapping performance with SP-SLAM\cite{zhang2019point}, PlanarSLAM\cite{li2021rgb}, and ManhattanSLAM\cite{yunus2021manhattanslam} is shown in Fig.\ref{plane_map} on sequences fr1-360, fr1-desk, fr1-room, fr2-desk, and fr3-lo. 
The Fig.\ref{plane_map} illustrates the performance of plane mapping on five sequences from the TUM dataset using SP-SLAM\cite{zhang2019point}, PlanarSLAM\cite{li2021rgb}, ManhattanSLAM\cite{yunus2021manhattanslam}, and our algorithm. In this context, ``w/o (C)(D)" indicates that our algorithm does not perform plane processing (Fig.\ref{framework}(C) and associates (Fig.\ref{framework}(D) solely through parameter comparisons. ``w/o (D)" signifies that our algorithm associates solely through parameter comparisons. These two experiments serve as ablations for Fig.\ref{framework}.(C) and (D). In Fig.\ref{plane_map}, different planes are represented using distinct colors.
 % For other sequences, either the scenes are almost identical to these five sequences but with different camera motions, or the scenes are too simple. Therefore, we have not provided comparative performance demonstrations for them. 
 It can be observed that SP-SLAM has a weaker perception of planes in the environment due to the lower granularity of point cloud selection when fitting planes with PCL. It neglects many planes in the scene. PlanarSLAM and ManhattanSLAM, like our algorithm, use the PEAC algorithm for plane extraction, which can extract more planes from the environment. However, PEAC struggle to ensure the quality of plane extraction. PlanarSLAM and ManhattanSLAM, lacking accurate and robust plane filtering and association strategies, seriously affect map accuracy. 
 % For example, in the fr3-lo sequence, PlanarSLAM fails to construct planes on objects like books and keyboards but creates numerous small planes on the teddy bear, further impacting localization accuracy. 
 When our method refrains from performing plane processing and relies solely on parameter-based comparisons for association, the constructed plane map is very messy due to the decreased quality of plane extraction and data association. when plane processing is carried out but association is solely based on parameter comparisons, planes extracted from objects and desktops are falsely associated, leading to the inability to achieve precise plane mapping. Our method benefits from accurate plane processing and association strategies, allowing us to construct accurate plane maps in planar ambiguous scenes. Our algorithm's performance in extracting plane vertices is shown in Fig.\ref{Ver_Extract}. It demonstrates the accuracy of our algorithm in perceiving structural information about planes.

\begin{table*}[t]
\begin{center}
\caption{Comparison of Absolute Trajectory Error (ATE) [m]. The best results are highlighted in bold and the second-best are underlined, and $/$ indicates localization failure.}\label{ATE-lable}
\renewcommand\arraystretch{1.2}
\setlength\tabcolsep{4.5pt}%调列距
\begin{tabular}{ccccccccccccccc}
\toprule
% \rule{0pt}{10pt} 
\specialrule{0em}{2pt}{1pt}
\multirow{2}{*}{Seq} & \multicolumn{2}{c}{ORB\cite{mur2017orb}}     & \multicolumn{2}{c}{SP\cite{zhang2019point}}    & \multicolumn{2}{c}{Planar\cite{li2021rgb}} & \multicolumn{2}{c}{Manhat\cite{yunus2021manhattanslam}} & \multicolumn{2}{c}{ w/o-(C)(D)} & \multicolumn{2}{c}{ w/o-(D)} & \multicolumn{2}{c}{Ours}          \\ 
% \rule{0pt}{1pt} 
\cline{2-15}
\specialrule{0em}{1pt}{2pt}
                           & RMSE            & S.D.            & RMSE            & S.D.           & RMSE      & S.D.            & RMSE           & S.D.          & RMSE        & S.D.  & RMSE           & S.D.          & RMSE        & S.D.                \\ \bottomrule 
fr1/360                    & 0.2516          & 0.1146          & \underline{0.1047}    & \underline{0.0377}    & 0.1068        & 0.0443        & 0.1297          & 0.0491          & 0.1178          & 0.0547       & 0.1537       & 0.0714       & \textbf{0.0830} & \textbf{0.0366} \\
fr1/desk                   & 0.0155          & 0.0088          & 0.0191          & 0.0086          & 0.1111        & 0.0428        & 0.0466          & 0.0227          & 0.0241          & 0.0131       & \underline{0.0153} & \underline{0.0082} & \textbf{0.0148} & \textbf{0.0079} \\
fr1/desk2                  & \underline{0.0215}    & \textbf{0.0092} & 0.0249          & 0.0137          & 0.2274        & 0.0944        & 0.0487          & 0.0177          & 0.0344          & 0.0161       & 0.0277       & 0.0140       & \textbf{0.0209} & \underline{0.0099}    \\
fr1/room                   & 0.0914          & \underline{0.0316}    & 0.0850          & 0.0340          & 0.1360        & 0.0710        & 0.1356          & 0.0504          & \underline{0.0814}    & 0.0419       & 0.0964       & 0.0331       & \textbf{0.0564} & \textbf{0.0153} \\
fr1/rpy                    & 0.0213          & 0.0129          & \underline{0.0204}    & \underline{0.0116}    & 0.4873        & 0.2504        & 0.0875          & 0.0225          & 0.0247          & 0.0150       & 0.0218       & 0.0132       & \textbf{0.0197} & \textbf{0.0115} \\
fr1/xyz                    & 0.0097          & \underline{0.0053}    & 0.0100          & 0.0054          & 0.0157        & 0.0084        & \textbf{0.0092} & \textbf{0.0046} & 0.0101          & 0.0054       & 0.0102       & 0.0055       & \underline{0.0095}    & 0.0053          \\
fr2/desk                   & \textbf{0.0114} & \underline{0.0043}    & 0.0224          & 0.0097          & 0.0386        & 0.0136        & 0.0332          & 0.0127          & 0.0236          & 0.0087       & 0.0289       & 0.0105       & \underline{0.0116}    & \textbf{0.0041} \\
fr2/xyz                    & \underline{0.0037}    & \textbf{0.0018} & 0.0043          & 0.0021          & 0.0148        & 0.0074        & 0.0086          & 0.0047          & 0.0056          & 0.0025       & 0.0039       & 0.0019       & \textbf{0.0036} & \underline{0.0019}    \\
fr2/rpy                    & \textbf{0.0031} & \textbf{0.0014} & \underline{0.0032}    & \underline{0.0014}    & 0.1355        & 0.0714        & 0.0040          & 0.0020          & 0.0047          & 0.0025       & 0.0035       & 0.0018       & 0.0038          & 0.0018          \\
fr3/cabinet                & /               & /               & \textbf{0.0151} & \textbf{0.0077} & 0.0477        & 0.0399        & 0.0228          & 0.0097          & \underline{0.0183}    & \underline{0.0091} & 0.0208       & 0.0108       & 0.0198          & 0.0095          \\
fr3/l-cabinet              & 0.0591          & 0.0262          & \textbf{0.0349} & \textbf{0.0183} & 0.2772        & 0.1397        & 0.0817          & 0.0338          & \underline{0.0455}    & \underline{0.0203} & 0.0750       & 0.0270       & 0.0731          & 0.0287          \\
fr3/loh                    & \underline{0.0136}    & \underline{0.0063}    & 0.0175          & 0.0073          & 0.1160        & 0.0820        & 0.0888          & 0.0505          & 0.0663          & 0.0236       & 0.0290       & 0.0106       & \textbf{0.0120} & \textbf{0.0052} \\
fr3/snf                    & /               & /               & 0.0226          & 0.0089          & 0.0212        & 0.0099        & 0.0406          & 0.0110          & 0.0265          & 0.0111       & \underline{0.0159} & \underline{0.0059} & \textbf{0.0157} & \textbf{0.0058} \\
fr3/snn                    & /               & /               & \underline{0.0125}    & \textbf{0.066}  & 0.0267        & 0.0094        & 0.0190          & 0.0093          & \textbf{0.0113} & \underline{0.0067} & 0.0163       & 0.0068       & 0.0166          & 0.0068          \\
fr3/stf                    & 0.0104          & 0.0045          & 0.0109          & 0.0046          & 0.0348        & 0.0119        & 0.0270          & 0.0061          & 0.0127          & 0.0042       & \underline{0.0096} & \underline{0.0041} & \textbf{0.0095} & \textbf{0.0040} \\
fr3/stn                    & \textbf{0.0106} & \textbf{0.0047} & 0.0297          & 0.0082          & 0.0547        & 0.0156        & \underline{0.0125}    & \underline{0.0067}    & 0.0266          & 0.0066       & 0.0263        & 0.0091       & 0.0282          & 0.0093                                   \\ 
\bottomrule
\end{tabular}
\end{center}
% \label{ATE-lable}
\vspace{-6mm} % 控制整个图（包括标题）与下面的距离
\end{table*}

%%%%%%%%%%%%%%%%%%%%%%%%

\subsection{Camera Localization Accuracy}
% 我们在TUM RGB-D数据集的16个序列上与ORB-SLAM2、SP-SLAM、PlanarSLAM和ManhattanSLAM的相机定位精度的对比结果如表*所示。其中“ w/o-CD”与“ w/o-D”是我们的消融实验算法，定义与上一节相同。
% 用于评估精度的指标是绝对轨迹误差(ATE)。ATE可以非常直观地反应算法精度和轨迹全局一致性。ATE的均方根误差(RMSE)和标准差误差(S.D.)用来表示系统的稳健性和稳定性。（The Root-Mean-Square-Error (RMSE) and Standard Deviation (S.D.) of both are used to represent the robustness and stability of the system\cite{fan2022blitz}.）所有的数据均来源于我们基于开源代码的运行结果，其中加粗数据表示最优结果，下横线数据表示次优结果，/表示定位失败。可以看到，我们的算法在绝大多数序列下取得了最优和次优的结果，这得益于准确的平面处理和关联策略，以及基于更多约束的优化策略。在平面结构较简单的序列下，例如fr2/rpy、fr3/cabinet、fr3/cabinet等，我们的算法优势较弱，这主要是由于在这样的场景下PEAC算法检测平面的精度不高且和平面相关的约束较少导致的。
The comparison results of camera localization accuracy, measured using Absolute Trajectory Error (ATE), on 16 sequences of the TUM RGB-D dataset with ORB-SLAM2\cite{mur2017orb}, SP-SLAM\cite{zhang2019point}, PlanarSLAM\cite{li2021rgb} and ManhattanSLAM\cite{yunus2021manhattanslam}, are presented in Table \ref{ATE-lable}. Where ``w/o (C)(D)" and ``w/o (D)" are our ablation study algorithms, defined the same as in the previous section. ATE provides an intuitive measure of algorithm accuracy and global trajectory consistency. The Root-Mean-Square-Error (RMSE) and Standard Deviation (S.D.) of both are used to represent the robustness and stability of the system. All data are based on our results using open-source code. It can be observed that our algorithm achieves the best and second-best results in the majority of sequences. This is attributed to accurate plane processing and association strategies, as well as optimization based on more constraints. In sequences with relatively simple plane structures, such as fr2/rpy, fr3/cabinet, fr3/cabinet, our algorithm exhibits weaker advantages, primarily due to the lower accuracy of plane detection using the PEAC algorithm and fewer constraints related to planes in such scenes.

%%%%%%%%%%%%%%%%%%%%%%%%
\subsection{Real-time Analysis}
% 为了完成对我们系统的整体评估，表*展示了我们的系统每个模块的平均耗时，表*展示了与ORB-SLAM2、SP-SLAM、PlanarSLAM和ManhattanSLAM的耗时对比结果。所有的测试均在fr1/desk序列上进行，表中的数据为每帧的平均耗时。在表*中OD表示目标检测模块，这部分与ORB特征提取并行运行；MDC表示漏检补偿模块；EPI表示平面和平面边缘点信息提取模块；PP表示图*中的平面处理模块；DA表示图*中的数据关联模块；UPM表示平面地图更新模块；MT表示完整的跟踪线程。可以看出，我们的算法的主要耗时集中在平面信息提取及处理、平面地图更新环节，跟踪线程的耗时与PlanarSLAM相当，优于ManhattanSLAM。
Table \ref{Real-time analysis} presents the average execution time for each module in our system, while Table \ref{Real-time analysis2} provides a comparison of execution times with ORB-SLAM2\cite{mur2017orb}, SP-SLAM\cite{zhang2019point}, PlanarSLAM\cite{li2021rgb} and ManhattanSLAM\cite{yunus2021manhattanslam}. All tests were conducted on the fr1/desk sequence, and the data in the tables represent the average time per frame. In Table \ref{Real-time analysis}, ``OD" represents the object detection module, which runs in parallel with ORB feature extraction. ``EPI" represents the module for extracting plane and plane edge point.``PP" corresponds to the plane processing module in Fig.\ref{framework}.(C). ``DA" represents the data association module in Fig.\ref{framework}.(D). ``UPM" stands for the plane map update module. Finally, ``MT" represents the complete tracking thread. 
% It can be observed that the main time consumption of our algorithm is concentrated in plane information extraction and processing, as well as plane map update steps. 
% The execution time of the tracking thread is comparable to PlanarSLAM and better than ManhattanSLAM.
Our algorithm's tracking thread has a total execution time comparable to PlanarSLAM and superior to ManhattanSLAM.

% \vspace{-2mm}

% ---->>>>>>>>>>>>> 表5 Real-time analysis
\begin{table}[h]%\tablefont
% \vspace{-2mm}
\begin{center}
\caption{THE AVERAGE RUNNING TIME OF EACH MODULE. [ms]}\label{Real-time analysis}
\setlength{\belowcaptionskip}{-0.5cm}   %调整图片标题与下文距离
\setlength{\tabcolsep}{1.5mm}{
\begin{tabular}{c|cccccc}
\hline
 Module & OD & EPI & PP & DA   & UPM & MT 
 \\ \hline
 Time  & 10.03        & 13.35            & 9.22         & 2.57 & 4.77           & 62.02 
 \\ \hline
\end{tabular}
}
\end{center}
\vspace{-3mm}
% \label{T-RPE-lable}
\end{table}

\vspace{-2mm}

\begin{table}[h]%\tablefont
% \vspace{-2mm}
\begin{center}
\caption{EXECUTION TIMES FOR DIFFERENT ALGORITHMS. [ms]}\label{Real-time analysis2}
\setlength{\belowcaptionskip}{-0.5cm}   %调整图片标题与下文距离
\setlength{\tabcolsep}{1.5mm}{
\begin{tabular}{c|ccccc}
\hline
 Algorithm & ORB\cite{mur2017orb} & SP\cite{zhang2019point} & Planar\cite{li2021rgb} & manhat\cite{yunus2021manhattanslam} & Ours 
 \\ \hline
 Time  & 17.71    & 24.07   & 59.91      & 92.35           & 62.02 
 \\ \hline
\end{tabular}
}
\end{center}
\vspace{-5mm}
% \label{T-RPE-lable}
\end{table}

\vspace{-2mm}

%%%%%%%%%%%%%%%%%%%%%%%%%%%%%%%%%%%%%%%%%%%%%%%%%%%%%%%%%%%%%%%%%%%%%%%%%%%%%%%%
\section{Conclusion} 
% 本文提出了一个完整的面向复杂场景的基于平面特征的视觉SLAM系统，包含平面处理、数据关联以及多约束因子图优化，旨在解决当前相关研究方案在复杂场景下定位和建图表现不佳的问题。广泛的实验表明，在复杂环境下，与其他最先进的方法相比，我们提出SLAM系统可以准确构建环境的平面地图，也一定程度上提升了相机定位精度。未来进一步的研究工作将集中在将该系统应用于增强现实与物体位姿估计。
This paper presents a comprehensive visual SLAM system based on plane features tailored for planar ambiguous scenes. The system encompasses plane processing, data association, and multi-constraint factor graph optimization, aiming to address the issues of poor localization and mapping performance in planar ambiguous scenes encountered by existing research solutions. Extensive experiments demonstrate that, in planar ambiguous scenes, our proposed SLAM system can accurately construct plane maps of the environment and also improve camera localization accuracy to a certain extent when compared to other state-of-the-art methods. Future research efforts will focus on applying this system to augmented reality and object pose estimation applications.

%\enlargethispage{-13.5cm}
%%%%%%%%%%%%%%%%%%%%%%%%%%%%%%%%%%%%%%%%%%%%%%%%%%%%%%%%%%%%%%%%%%%%%%%%%%%%%%%%

\bibliographystyle{IEEEtran}
\bibliography{PCS-SLAM}

%%%%%%%%%%%%%%%%%%%%%%%%%%%%%%%%%%%%%%%%%%%%%%%%%%%%%%%%%%%%%%%%%%%%%%%%%%%%%%%%

%%%%%%%%%%%%%%%%%%%%%%%%%%%%%%%%%%%%%%%%%%%%%%%%%%%%%%%%%%%%%%%%%%%%%%%%%%%%%%%%

\end{document}